\documentclass[11pt]{article}

\usepackage{colortbl}

\usepackage[]{acl}

\usepackage{times}
\usepackage{latexsym}

\usepackage[T1]{fontenc}

\usepackage[utf8]{inputenc}

\usepackage{microtype}

\usepackage{textcomp}

\usepackage{graphicx}

\usepackage{algorithm}
\usepackage{caption}
\usepackage[noend]{algpseudocode}
\usepackage{colortbl}
\usepackage{xcolor}
\usepackage{hyperref}
\usepackage{multirow}
\usepackage{multicol}
\usepackage{CJKutf8}
\usepackage{textcomp}

\usepackage{booktabs}
\newcommand{\scell}[1]{\multicolumn{1}{r|}{\scriptsize{#1}}}
\newcommand{\srcell}[1]{\multicolumn{1}{r}{\scriptsize{#1}}}
\newcommand{\sgcell}[1]{\multicolumn{1}{r|}{\cellcolor{Gray}{\scriptsize{#1}}}}
\newcommand{\sgrcell}[1]{\multicolumn{1}{r}{\cellcolor{Gray}{\scriptsize{#1}}}}

\definecolor{Gray}{gray}{0.9}
\newcolumntype{a}{>{\columncolor{Gray}}c}

\usepackage{setspace}

\title{EAG: Extract and Generate Multi-way Aligned Corpus for Complete Multi-lingual Neural Machine Translation}

\author{
  Yulin Xu\textsuperscript{1}\thanks{ \ \ Equal contribution. Work was done when Yulin Xu was interning at Pattern Recognition Center, WeChat AI, Tencent Inc, China.}   , 
  Zhen Yang\textsuperscript{1}$^{*}$\thanks{ \ \ Corresponding Author.}, 
  \textbf{Fandong Meng}\textsuperscript{1}, 
  and \textbf{Jie Zhou}\textsuperscript{1}\\
  \textsuperscript{1}Pattern Recognition Center, WeChat AI, Tencent Inc, China \\
  \texttt{\{xuyulincs\}@gmail.com} \\
  \texttt{\{zieenyang,fandongmeng,withtomzhou\}@tencent.com} \\
}

\begin{document}
\begin{CJK}{UTF8}{gbsn}
\maketitle
\begin{abstract}
Complete Multi-lingual Neural Machine Translation (C-MNMT) achieves superior performance against the conventional MNMT by constructing multi-way aligned corpus, i.e., aligning bilingual training examples from different language pairs when either their source or target sides are identical. However, since exactly identical sentences from different language pairs are scarce, the power of the multi-way aligned corpus is limited by its scale. To handle this problem, this paper proposes "Extract and Generate" (EAG), a two-step approach to construct large-scale and high-quality multi-way aligned corpus from bilingual data. Specifically, we first extract candidate aligned examples by pairing the bilingual examples from different language pairs with highly similar source or target sentences; and then generate the final aligned examples from the candidates with a well-trained generation model. With this two-step pipeline, EAG can construct a large-scale and multi-way aligned corpus whose diversity is almost identical to the original bilingual corpus. Experiments on two publicly available datasets i.e., WMT-5 and OPUS-100, show that the proposed method achieves significant improvements over strong baselines, with +1.1 and +1.4 BLEU points improvements on the two datasets respectively.
\end{abstract}

\section{Introduction} \label{sec:intro}




Multilingual Neural Machine Translation (MMMT) \cite{dong2015multi,firat2017multi,johnson2017google,aharoni2019massively} has achieved promising results on serving translations between multiple language pairs with one model. With sharing parameters of the model, MNMT can facilitate information sharing between similar languages and make it possible to translate between low-resource and zero-shot language pairs. Since the majority of available MT training data are English-centric, i.e., English either as the source or target language, most non-English language pairs do not see a single training example when training MNMT models \cite{freitag2020complete}. Therefore, the performance of MNMT models on non-English translation directions still left much to be desired: 1) Lack of training data leads to lower performance for non-English language pairs\cite{zhang-etal-2021-competence-based}; 2) MNMT models cannot beat the pivot-based baseline systems which translate non-English language pairs by bridging through English \cite{cheng2016neural,habash2009improving}.

Recently, \citet{freitag2020complete} re-kindle the flame by proposing C-MNMT, which trains the model on the constructed multi-way aligned corpus. Specifically, they extract the multi-way aligned examples by aligning training examples from different language pairs when either their source or target sides are identical (i.e., pivoting through English, for German$\rightarrow$English and English$\rightarrow$French to extract German-French-English examples). Since they directly extract the multi-way aligned examples from the bilingual corpus, we refer to their approach as the \emph{extraction-based} approach. Despite improving the performance,  the scale of multi-way aligned corpus extracted by \citet{freitag2020complete} is always limited compared to English-centric bilingual corpus, e.g., only 0.3M German-Russian-English multi-way aligned corpus extracted from 4.5M German-English and 33.5M English-Russian bilingual corpus. A simple idea for remedying this problem is to add the roughly-aligned corpus by extracting the training examples when either their source or target sides are highly similar. However, our preliminary experiments show that the performance of the model decreases dramatically when we train the model with appending the roughly-aligned corpus.\footnote{Detailed descriptions about the preliminary experiment can be found in Section \ref{ana:simi}.} One possible solution, referred to as the \emph{generation-based} approach, is to generate the multi-way aligned examples by distilling the knowledge of the existing NMT model, e.g., extracting German-English-French synthetic three-way aligned data by feeding the English-side sentences of German-English bilingual corpus into the English-French translation model. Although the \emph{generation-based} approach can theoretically generate non-English corpus with the same size as original bilingual corpus, its generated corpus has very low diversity as the search space of the beam search used by NMT is too narrow to extract diverse translations \cite{wu2020generating,sun2020generating,shen2019mixture}, which severely limits the power of the \emph{generation-based} approach.


In order to combine advantages of the two branches of approaches mentioned above, we propose a novel two-step approach, named EAG (Extract and Generate), to construct large-scale and high-quality multi-way aligned corpus for C-MNMT. Specifically, we first extract candidate aligned training examples from different language pairs when either their source or target sides are highly similar; and then we generate the final aligned examples from the pre-extracted candidates with a well-trained generation model.
The motivation behind EAG is two-fold: 1) Although identical source or target sentences between bilingual examples from different language pairs are scarce, highly similar sentences in source or target side are more wide-spread; 2) Based on the pre-extracted candidate aligned examples which have highly similar source or target sentences, EAG can generate the final aligned examples by only refining the sentences partly with a few modifications. Therefore, the non-English corpus constructed by EAG has almost identical diversity to the original bilingual corpus. Experiments on the publicly available data sets, i.e., WMT-5 and OPUS-100, show that the proposed method achieves substantial improvements over strong baselines. 

\section{Background} \label{sec:background}

\paragraph{Bilingual NMT} 
Neural machine translation \cite{2014Sequence,cho2014learning,vaswani2017attention} achieves great success in recent years due to its end-to-end learning approach and large-scale bilingual corpus. Given a set of sentence pairs $D=\left \{ (x, y) \in (X \times Y) \right \}$, the NMT model is trained to learn the parameter $\theta$ by maximizing the log-likelihood $\sum\nolimits_{(x,y) \in D}log P(y|x; \theta)$. 
\paragraph{MNMT}
Considering training a separate model for each language pair is resource consuming, MNMT \cite{dong2015multi,johnson2017google,gu2020improved} is introduced to translate between multiple language pairs using a single model \cite{johnson2017google,ha2toward,lakew2018comparison}. We mainly focus on the mainstream MNMT model proposed by \citet{johnson2017google}, which only introduces an artificial token to the input sequence to indicate which target language to translate.
\paragraph{C-MNMT}
C-MNMT is proposed to build a complete translation graph for MNMT, which contains training examples for each language pair \cite{freitag2020complete}. 
A challenging task remaining is how to get direct training data for non-English language pairs. In \citet{freitag2020complete}, non-English training examples are constructed by pairing the non-English sides of two training examples with identical English sides. However, this method can't get large-scale training examples since the quantity of exactly identical English sentences from different language pairs is small. Another feasible solution is to generate training examples with pivot-based translation where the source sentence cascades through the pre-trained source $\rightarrow$ English and English $\rightarrow$ target systems to generate the target sentence \cite{cheng2016neural}. Despite a large quantity of corpus it can generate, its generated corpus has very low diversity \cite{wu2020generating,sun2020generating,shen2019mixture}. 

\section{Methods} \label{sec: methods}
The proposed EAG has a two-step pipeline. The first step is to extract the candidate aligned examples from the English-centric bilingual corpus. The second step is to generate the final aligned examples from the candidates extracted in the first step.

\begin{figure*}[t]
    \centering
    \includegraphics[scale=0.53]{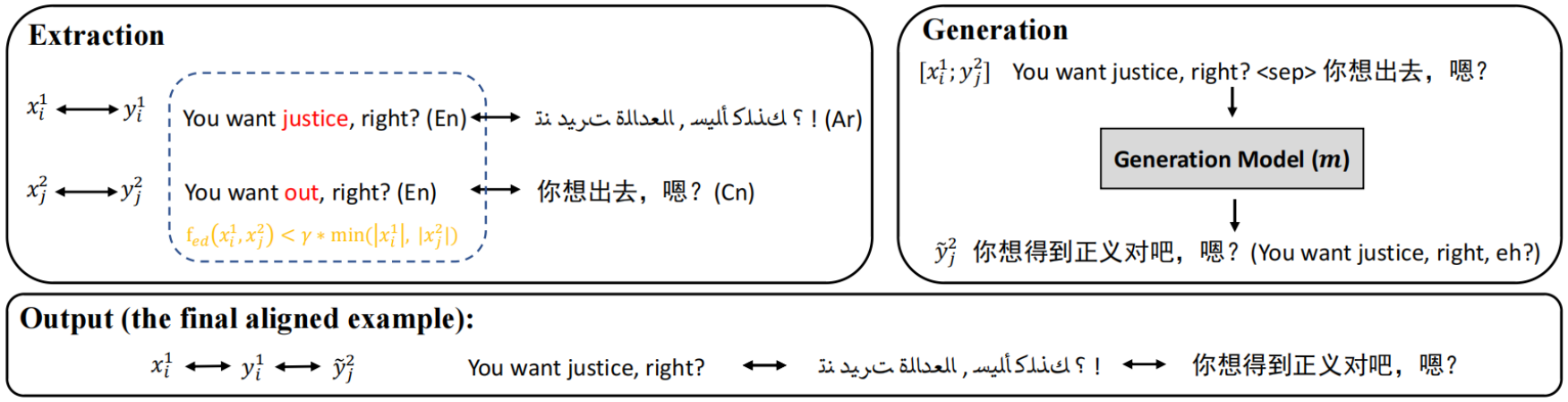}
    \caption {Examples constructed by EAG. "$x^1_i\leftrightarrow y^1_i$" and "$x^2_j \leftrightarrow y^2_j$" represent the bilingual examples in English $\rightarrow$ Arabic and English $\rightarrow$ Chinese respectively. $\tilde{y}^2_j$ is the generated Chinese sentence, which is aligned to $x^1_i$ and $y^1_i$. For a clear presentation, the Google translation (in English) for the generated $\tilde{y}_j^2$ is also provided.}
    \label{fig:toy example}
    \vspace{-0.2cm}
\end{figure*}

\subsection{Extract candidate aligned examples} 
Different from \citet{freitag2020complete} who extract non-English training examples by aligning the English-centric bilingual training examples with identical English sentences, we extract the candidate aligned examples by pairing two English-centric training examples with highly similar English sentences. Various metrics have been proposed to measure the superficial similarity of two sentences, such as TF-IDF \cite{aizawa2003information,huang2011text}, edit distance \cite{xiao2008ed,deng2013top}, etc. In this paper, we take edit distance as the measurement to decide the superficial similarity of two English sentences.
Three main considerations are behind. Firstly, since edit distance measures the similarity of two sentences with the minimum number of operations to transform one into the other, it tends to extract sentences with similar word compositions and sentence structures. Secondly, since edit distance only utilizes three operations, i.e., removal, insertion, or substitution, it is easier to mimic these operations in the process of generating the final aligned examples (we leave the explanation in the next subsection). Finally, unlike TF-IDF which only considers word bags in two sentences, edit distance also considers the word order in each sentence.

Formally, given two English-centric bilingual corpora from two different language pairs $\{X^1, Y^1\}$ and $\{X^2, Y^2\}$, where $X^1$ and $X^2$ are English sides, $Y^1$ and $Y^2$ belong to language $L_a$ and $L_b$ respectively. For sentence pair $(x^1_i, y^1_i)\in \{X^1, Y^1\}$ and $(x^2_j, y^2_j)\in \{X^2, Y^2\}$, we take $(x^1_i, y^1_i, x^2_j, y^2_j)$ as a candidate aligned example if the two English sentences $x^1_i$ and $x^2_j$ meets:
\begin{equation}
 f_{ed}(x^1_i, x^2_j) \leq \gamma * min(|x^1_i|, |x^2_j|) , \gamma \in (0,1)
\end{equation}
where $f_{ed}$ refers to the function of edit distance calculation, $|x|$ represents the length of the sentence $x$, $\gamma$ is the similarity threshold which can be set by users beforehand to control the similarity of sentences in the candidate aligned examples. With setting $\gamma=0$, we can directly extract the same multi-way aligned examples with \citet{freitag2020complete}.
With larger $\gamma$, more candidate aligned examples can be extracted for looser restriction. Accordingly, there are more noises in the extracted candidate aligned examples.

\subsection{Generate final aligned examples} 
\label{sec:gene_final}
In the extracted candidate aligned example $(x^1_i, y^1_i, x^2_j, y^2_j)$, $(x^2_j, y^2_j)$ is not well aligned to $(x^1_i, y^1_i)$ if $f_{ed}(x^1_i, x^2_j)$ does not equal to zero. To construct the final three-way aligned example, we search for one sentence pair $(\tilde{x}^2_j, \tilde{y}^2_j)$ in the language pair $\{X^2, Y^2\}$, where $\tilde{x}^2_j$ has the same meaning to $x^1_i$ (thus $(x^1_i, y^1_i,\tilde{y}^2_j)$ is a three-way aligned example). Unfortunately, it is very difficult for us to directly find such a sentence pair in the large search space. However, considering $\tilde{x}^2_j$ and $x^1_i$ are both in English, we can take an extreme case where $\tilde{x}^2_j$ is identical to $x^1_i$ in the superficial form. Now, the remained question is that we need to search for the sentence $\tilde{y}^2_j$ in language $L_b$, which has the same meaning to $x^1_i$. By comparing $(x^1_i, \tilde{y}^2_j)$ with $(x^2_j, y^2_j)$, as $x^1_i$ can be transformed from $x^2_j$ with the operations performed by edit distance, it is naturally to suppose that we can find such a $\tilde{y}^2_j$ which can be transformed from $y^2_j$ with these operations similarly. Therefore, we can limit the search space for $\tilde{y}^2_j$ with two restrictions: Firstly, sentence $\tilde{y}^2_j$ has the same meaning with $x^1_i$; Secondly, $\tilde{y}^2_j$ is transformed from $y^2_j$ with the operations performed by edit distance. Considering the restrictions mentioned above, we apply an NMT model $m$ to search and generate $\tilde{y}^2_j$. There are two main questions left to be resolved: how to train such a model $m$ and how to generate $\tilde{y}^2_j$ with a well-trained $m$.

\paragraph{Training} 
Motivated by the recent success of self-supervised training \cite{devlin2018bert,conneau2019cross,song2019mass,yang2020csp} in natural language processing, we automatically construct the training corpus for $m$ from the candidate aligned examples. Given the candidate aligned example $(x^1_i, y^1_i, x^2_j, y^2_j)$, the training example for $m$ is built as:
 \begin{equation}
  ([x^2_j;\hat{y}^2_j], y^2_j)
 \end{equation}
 where $y^2_j$ is the target sentence, the concatenation of $x^2_j$ and $\hat{y}^2_j$ is the source-side input. $\hat{y}^2_j$ is the noisy form of $y^2_j$ which we build by mimicking the operations of edit distance, i.e, performing insertion, removal, or substitution on some pieces of $y^2_j$ randomly. Specifically, with probability $\beta$, each position of sentence $y^2_j$ can be noised by either removed directly, inserted or substituted with any other words in the dictionary $W_b$, which is constructed from the corpus $Y^2$. With the self-constructed training examples, the model $m$ is trained to generate the target sentence, which is recovered from the right-side of the concatenated input with the operations performed by edit distance, and has the same meaning to the left-side of the input.
\paragraph{Generating} With a well-trained $m$, we generate the final aligned examples by running the inference step of $m$. Formally, for the final aligned example $(x^1_i, y^1_i, \tilde{y}^2_j)$, the sentence $\tilde{y}^2_j$ is calculated by:
 \begin{equation}
  \tilde{y}^2_j = m([x^1_i;y^2_j])
 \end{equation}
 where $[\cdot;\cdot]$ represents the operation of concatenation, and $m(x)$ refers to running the inference step of $m$ with $x$ fed as input. With this generation process, $\tilde{y}^2_j$ is not only has the same meaning to $x^1_i$ (thus also aligned to $y^1_i$), but also keeps the word composition and sentence structure similar to $y^2_j$. Therefore, EAG can construct the final aligned corpus for each non-English language pair, and keep the diversity of the constructed corpus almost identical to the original English-centric corpus. For a clear presentation,  Algorithm \ref{alg:training} in Appendix \ref{app:algorithm} summarizes the process of generating the final aligned examples. We also provide a toy example in Figure \ref{fig:toy example} to illustrate how the proposed EAG works.

\section{Experiments and Results} \label{sec: experiment}
For fair comparison, we evaluate our methods on the publicly available dataset WMT-5, which is used by \citet{freitag2020complete}.
Additionally, we test the scalability of our method by further conducting experiments on Opus-100, which contains English-centric bilingual data from 100 language pairs \cite{zhang2020improving}.
In the extraction process, we run our extraction code on the CPU with 24 cores and 200G memory.\footnote{Intel(R) Xeon(R) Platinum 8255C CPU @ 2.50GHz} In the generation process,  we take transformer-big \cite{vaswani2017attention} as the configuration for $m$, and $m$ is trained with the self-constructed examples mentioned in Section \ref{sec:gene_final} on eight V100 GPU cards.\footnote{Detained training process for $m$ can be found in the Appendix \ref{app:model}.}

We choose Transformer as the basic structure for our model and conduct experiments on two standard configurations, i.e, transformer-base and transformer-big. All models are implemented based on the open-source toolkit fairseq 
\cite{ott2019fairseq} and trained on the machine with eight V100 GPU cards.\footnote{We upload the code as supplementary material for review, and it will be released publicly upon publication.} All bilingual models are trained for 300,000 steps and multi-lingual models are trained for 500,000 steps. We add a language token at the beginning of the input sentence to specify the required target language for all of the multi-lingual models. For the hyper-parameters $\beta$ and $\gamma$, we set them as 0.5 and 0.3 by default and also investigate how their values produce effects on the translation performance.
\subsection{Experiments on WMT-5}
\subsubsection{Datasets and pre-processing}
Following \citet{freitag2020complete}, we take WMT13EnEs, WMT14EnDe, WMT15EnFr, WMT18EnCs and WMT18EnRu as the training data, the multi-way test set released by WMT2013 evaluation campaign \cite{bojar2014findings} as the test set. The size of each bilingual training corpus (the non-English corpus constructed by \citet{freitag2020complete} included) is presented in Table \ref{tab:wmt_training_data_stats}. For the bilingual translation task, the source and target languages are jointly tokenized into 32,000 sub-word units with BPE \cite{sennrich-etal-2016-neural}. The multi-lingual models use a vocabulary of 64,000 sub-word units tokenized from the combination of all the training corpus. Similar to  \citet{freitag2020complete}, we use a temperature-based data sampling strategy to over-sample low-resource language pairs in standard MNMT models and low-resource target-languages in C-MNMT models (temperature $T=5$ for both cases). We use BLEU scores \cite{papineni2002bleu} to measure the model performance and all BLEU scores are calculated with sacreBLEU \cite{post-2018-call}.\footnote{sacreBLEU signatures: BLEU+case.mixed+lang.SRC-TGT+numrefs.1+smooth.exp+tok.intl+version.1.5.1}
\begin{table}[ht]
\begin{center}
{\setlength{\tabcolsep}{.35em}
\begin{tabular}{l||c|c|a|c|c|c}
 & cs & de &  en & es & fr & ru\\ \hline \hline
cs &  & 0.7 & 47 & 0.8 & 1 & 0.9 \\ \hline
de & 0.7 &  & 4.5 & 2.3 & 2.5 & 0.3 \\ \hline
\rowcolor{Gray}
en & 47 & 4.5 & & 13.1 & 38.1 &  33.5 \\ \hline
es & 0.8 & 2.3 & 13.1 &  & 10 & 4.4 \\ \hline
fr & 1 & 2.5 & 38.1 & 10 &  & 4.8 \\ \hline
ru & 0.9 & 0.3 & 33.5 & 4.4 & 4.8 &  \\
\end{tabular}
}
\end{center}
\vspace{-0.5em}
\caption{WMT: Available training data (in million) after constructing non-English examples by \cite{freitag2020complete}.}
\label{tab:wmt_training_data_stats}
\end{table}
\begin{table}[ht]
\begin{center}
{\setlength{\tabcolsep}{.35em}
\begin{tabular}{l||c|c|a|c|c|c}
   & cs & de & en & es & fr & ru\\ \hline \hline
   cs &  & 2.2 & 47 & 2.5 & 4.1 & 3.2 \\ \hline
   de & 2.2 &  & 4.5 & 7.1 & 6.1 & 1.4 \\ \hline
   \rowcolor{Gray}
   en & 47 & 4.5 &  & 13.1 & 38.1 & 33.5 \\ \hline
   es & 2.5 & 7.1 & 13.1 &  & 22 & 10.1 \\ \hline
   fr & 4.1 & 6.1 & 38.1 & 22 &  & 11.0 \\ \hline
   ru & 3.2 & 1.4 & 33.5 & 10.1 & 11.0 &  \\
\end{tabular}
}
\end{center}
\vspace{-0.5em}
\caption{WMT: Available training data (in million) after constructing non-English examples by EAG.}
\label{tab:wmt_training_data_stats_EAG}
\end{table}

\subsubsection{Corpus constructed by EAG}
Table \ref{tab:wmt_training_data_stats_EAG} shows the training data after constructing non-English examples from English-centric corpus by the proposed EAG. By comparing Table \ref{tab:wmt_training_data_stats_EAG} with Table \ref{tab:wmt_training_data_stats}, we can find that EAG can construct much more multi-way aligned non-English training examples than \citet{freitag2020complete}, e.g., EAG constructs 1.4M bilingual training corpus for the language pair German $\rightarrow$ Russian which is almost up to 4 times more than the corpus extracted by \citet{freitag2020complete}. In all, EAG constructs no less than 1M bilingual training examples for each non-English language pair.

\begin{table*}[ht]
\begin{center}
{\setlength{\tabcolsep}{.16em}
\scalebox{0.85}{
\begin{tabular}{lcccccccc}
	\toprule[2pt]
		System & En-X & De-X & Fr-X & Ru-X & Es-X & Cs-X & English-centric & non-English\\
		\midrule[1pt]
		bilingual system \cite{vaswani2017attention} & 28.8 & 22.4 & 24.6 & 19.6 & 25.9& 21.1 & 30.3 & 20.4 \\
        MNMT system \cite{johnson2017google} & 28.6& 16& 15.4& 14.2 & 15.0& 20.3& 30.0 & 12.3 \\
        pivot system \cite{cheng2016neural} & 28.8 & 25.7& 26.1& 25.2 &27.4 &26.4 & 30.3 & 24.7 \\
        generation-based C-MNMT& 28.8 & 26.2 & 26.7 & 26.0 & 27.5 &26.8 & 30.3 & 25.3 \\ 
        extraction-based C-MNMT \cite{freitag2020complete}& 29.3& 27.2& 27.3 & 26.7 & 28.6 &28.2 & 30.6 & 26.8 \\
        \midrule[1pt]
        \textbf{EAG} & \textbf{29.6$^{*}$} & \textbf{28.3$^{*}$}& \textbf{28.2$^{*}$}& \textbf{27.7$^{*}$}& \textbf{29.5$^{*}$}& \textbf{29.7$^{*}$}& \textbf{30.8} & \textbf{27.9$^{*}$} \\
		\bottomrule[2pt]
	       \end{tabular}
}
}
\vspace{-0.5em}
\end{center}
\caption{
The translation performance for different systems on WMT data. '$L$-X' means the set of translation directions from language $L$  to other five languages. 'English-centric' and 'non-English' refer to the set for English-centirc and non-English language pairs respectively. For each set, bold indicates the highest value, and $*$ means the gains are statistically significant with $p < 0.05$ compared with extraction-based C-MNMT.
}
\label{tab:Result_SNMT}
\vspace{-1em}
\end{table*}

\subsubsection{Baselines}
In order to properly and thoughtfully evaluate the proposed method, we take the following five kinds of baseline systems for comparison:
\paragraph{Bilingual systems \cite{vaswani2017attention}}
Apart from training bilingual baseline models on the original English-centric WMT data, we also train bilingual models for non-English language pairs on the direct bilingual examples extracted by \citet{freitag2020complete}.
\paragraph{Standard MNMT systems \cite{johnson2017google}}
We train a standard multi-lingual NMT model on the original English-centric WMT data. 
\paragraph{Bridging (pivoting) systems \cite{cheng2016neural}} 
In the bridging or pivoting system, the source sentence cascades through the pre-trained source $\rightarrow$ English and English $\rightarrow$ target systems to generate the target sentence.

\paragraph{Extraction-based C-MNMT systems \cite{freitag2020complete}}
\citet{freitag2020complete} construct the multi-way aligned examples by directly extracting and pairing bilingual examples from different language pairs with identical English sentences. 


\paragraph{Generation-based C-MNMT systems} The generation-based C-MNMT baselines construct non-English bilingual examples by distilling the knowledge of the system which cascades the source $\rightarrow$ English and English $\rightarrow$ target models. Different from the bridging baselines which just feed the test examples into the cascaded system and then measure the performance on the test examples, the generation-based C-MNMT baselines feed the non-English sides of the bilingual training examples into the cascaded systems and then get the non-English bilingual training examples by pairing the inputs and outputs. The combination of the generated non-English corpus and original English-centric corpus is used to train the C-MNMT model.
\subsubsection{Results}
We first report the results of our implementations and then present the comparisons with previous works. In our implementations, we take the transformer-base as the basic model structure since it takes less time and computing resources for training. To make a fair comparison with previous works, we conduct experiments on transformer-big which is used by baseline models.
\paragraph{Results of our implementation}
Table \ref{tab:Result_SNMT} shows the results of our implemented systems. Apart from the average performance of the translation directions from each language to others, we also report the average performance on the English-centric and non-English language pairs.\footnote{Readers can find the detailed results for each language pair in the Appendix \ref{app:bleu}.} As shown in Table \ref{tab:Result_SNMT}, we can find that the proposed EAG achieves better performance than all of the baseline systems. Compared to the extraction-based C-MNMT, the proposed method achieves an improvement up to 1.1 BLEU points on non-English language pairs. The generation-based C-MNMT performs worse than the extraction-based one even if it generates much larger corpus. Since there is no any training example for non-English language pairs in standard MNMT, the standard MNMT system achieves inferior performance to the pivot and bilingual systems on non-English translation directions. However, with the constructed non-English training examples, EAG achieves 3.2 and 7.5 BLEU points improvements compared with the pivot and bilingual systems respectively. 
\begin{table}[ht]
\begin{flushleft}
{\setlength{\tabcolsep}{.39em}
\begin{tabular}{ cl||c|c|a|c|c|c}
\multicolumn{2}{c}{~} & \multicolumn{6}{c}{target} \\
\multirow{12}{*}{\rotatebox[origin=c]{90}{source}} & & cs & de & \cellcolor{white}{en} & es & fr & ru\\ \cline{2-8} \cline{2-8}
& cs & & 27.6 & 31.9 & 31.6 & 33.8 & 28.4 \\ [-0.25em] 
& & & \scell{+1.8} & \sgcell{-0.1} & \scell{+1.5} & \scell{+2.4} & \srcell{+1.5} \\ \cline{2-8}
& de & 25.8 & & 31.4 & 31.2 & 33.3 & 25.1 \\ [-0.25em] 
& & \scell{+1.9} & & \sgcell{+0.2} & \scell{+1.3} &  \scell{+1.5} & \srcell{+1.7} \\ \cline{2-8}
& en & \cellcolor{Gray}{26.7} & \cellcolor{Gray}{27.4} & & \cellcolor{Gray}{35.1} & \cellcolor{Gray}{36.0} & \cellcolor{Gray}{26.6}  \\ [-0.25em] 
& & \sgcell{-0.2} & \sgcell{+0.3} & & \sgcell{+0.1} & \sgcell{+0.5} & \sgrcell{+0.2}\\ \cline{2-8}
& es & 25.9 & 26.4 & 35.1 & & 36.8 & 25.6 \\ [-0.25em]
& & \scell{+1.0} & \scell{+0.7} & \sgcell{+0.2} & & \scell{+0.8} & \srcell{+0.7} \\ \cline{2-8} 
& fr & 24.9 & 26.5 & 34.6 & 33.8 & & 23.7 \\ [-0.25em] 
& & \scell{+1.2} & \scell{+1.3} & \sgcell{+0.2} & \scell{+0.5} & & \srcell{+0.2} \\ \cline{2-8}
& ru & 24.9 & 25.1 & 30.1 & 30.4 & 31.0 & \\ [-0.25em]
& & \scell{+0.6} & \scell{+2.4} & \sgcell{+0.3} & \scell{+1.8} & \scell{+0.9} & \\
\end{tabular}
}
\end{flushleft}
\vspace{-0.5em}
\caption{Results for transformer-big trained on corpus constructed by EAG. The small numbers are the difference with respect to \citet{freitag2020complete}.}
\label{tab:big_result_EAG}
\end{table}
\paragraph{Results compared with previous works}
Table \ref{tab:big_result_EAG} shows the results of the proposed EAG. We can find that the proposed EAG surpasses \citet{freitag2020complete} almost on all of the translation directions, and achieves an improvement with up to 2.4 BLEU points on the Russian-to-German direction.
\subsection{Experiments on Opus-100}
\paragraph{Datasets and pre-processing}
\citet{zhang2020improving} first create the corpus of Opus-100 by sampling from the OPUS collection \cite{tiedemann2012parallel}. Opus-100 is an English-centric dataset which contains 100 languages on both sides and up to 1M training pairs for each language pair. To evaluate the performance of non-English language pairs, \citet{zhang2020improving} sample 2000 sentence pairs of test data for each of the 15 pairings of Arabic, Chinese, Dutch, French, German, and Russian. Following \citet{zhang2020improving}, we report the sacreBLEU on the average of the 15 non-English language pairs.\footnote{Signature: BLEU+case.mixed+numrefs.1+smooth.exp+
tok.13a+version.1.4.1} The statistics about the non-English corpus constructed by \citet{freitag2020complete} and EAG
are presented in Table \ref{tab:statics_opus}. 
We can find that EAG is able to construct much more bilingual corpus for non-English language pairs (almost nine times more than \citet{freitag2020complete} for each language pair). We use a vocabulary of 64,000 sub-word units for all of the multi-lingual models, which is
tokenized from the combination of all the training corpus with SentencePiece. 

\begin{table}[ht]
\centering
 \scalebox{0.90}{
	\begin{tabular}{lcc}
	\toprule[2pt]
	& \citet{freitag2020complete} & EAG \\
    \midrule[1pt]
		Ar-X & 0.19 & 1.12 \\
		De-X & 0.14 & 1.01 \\
		Fr-X & 0.15 &1.25 \\
		Cn-X & 0.16 & 1.03 \\
		Ru-X & 0.18 & 0.94 \\
		Nl-X & 0.18 & 0.98 \\
	\bottomrule[2pt]
	\end{tabular}
	}
\caption{\label{tab:statics_opus} The amount of non-English examples (in million) constructed from bilingual examples in OPUS-100. "$L$-X" means the total number of the corpus for the directions from language $L$ to other five.}
\end{table}

\begin{table}[ht]
\centering
 \scalebox{0.85}{
\begin{tabular}{lc}
	\toprule[2pt]
	System  & non-English \\
	\midrule[1pt]
        MNMT system &  4.5 \\
        pivot system &  13.1\\
        generation-based C-MNMT &  13.8\\
        extraction-based C-MNMT &  16.5\\
        \midrule[1pt]
        \citet{zhang2020improving} & 14.1\\
        \citet{fan2020beyond} & \textbf{18.4} \\
        \midrule[1pt]
        \textbf{EAG} &  \ 17.9$^{*}$ \\
		\bottomrule[2pt]
	    \end{tabular}}
	    \caption{\label{tab:Result_OPUS} Results on Opus-100. We directly cite their results for \citet{zhang2020improving} and \citet{fan2020beyond}. ${*}$ means the gains of EAG  are significant compared with extraction-based C-MNMT ($p < 0.05$)}
\end{table}
\vspace{-0.2cm}

\paragraph{Results}
Apart from the baselines mentioned above, we also compare with other two systems proposed by \citet{zhang2020improving} and \citet{fan2020beyond}. \citet{zhang2020improving} propose the online back-translation for improving the performance of non-English language pairs in Opus-100. \citet{fan2020beyond} build a C-MNMT model, named ${m2m}_{100}$, which is trained on 7.5B training examples built in house. Following \citet{zhang2020improving}, we take the transformer-base as the basic model structure for the experiments and results are reported in Table \ref{tab:Result_OPUS}. We can find that EAG achieves comparable performance to \citet{fan2020beyond} which utilizes much more data than ours. This is not a fair comparison as the data used \citet{fan2020beyond} is 75 times as much as ours. Additionally, our model surpasses all other baseline systems and achieves +1.4 BLEU points improvement compared with the extraction-based C-MNMT model. 

\section{Analysis} \label{sec: analysis}
We analyze the proposed method on Opus-100 and take the transformer-base as the model structure.

\subsection{Effects of the hyper-parameters}
\label{ana:simi}
\begin{figure}[htb]
   \begin{center}
   \includegraphics[scale=0.39]{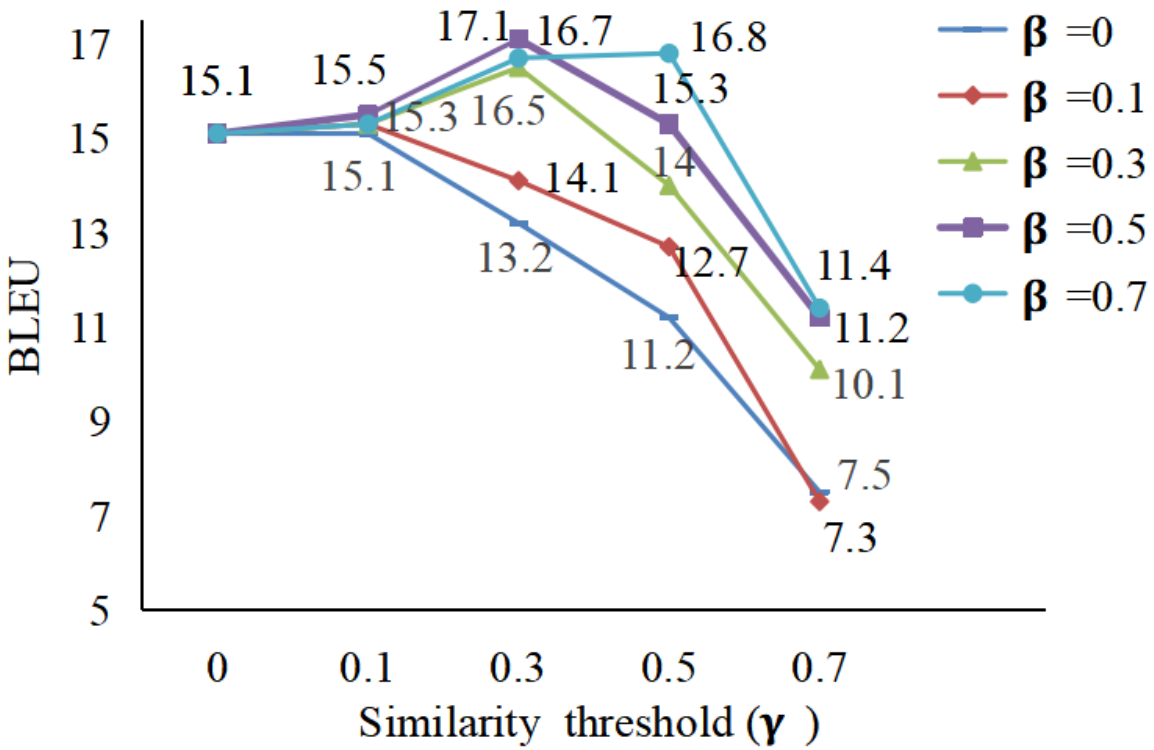}
   \end{center}
   \caption{\label{fig:gama} Experimental results on testing the hyper-parameters. With $\gamma$ set as 0, this is the setting of the baseline system of \citet{freitag2020complete}.}	
\end{figure}

The similarity threshold $\gamma$ and the noise ratio $\beta$ are important hyper-parameters in EAG. In this section, we want to test how these two hyper-parameters affect the final translation performance and how they work with each other. We investigate this problem by studying the translation performance with different $\gamma$ and $\beta$, where we vary $\gamma$ and $\beta$ from 0 to 0.7 with the interval 0.2. We report the average BLEU score for the translation directions from Arabic to other five languages on the development sets built in house. Figure \ref{fig:gama} shows the experimental results. With $\beta=0$, it means that the generation process is not applied and we directly train the NMT model with the extracted roughly aligned examples. And this is the setting of our motivated experiments mentioned in Section \ref{sec:intro}. We can find that, the final performance drops sharply when we directly train the model with the roughly aligned sentence pairs. For each curve in Figure \ref{fig:gama}, we can find that the model achieves the best performance when the $\gamma$ is around $\beta$, and then the performance decreases with $\gamma$ growing. A relatively unexpected result is that the model usually achieves the best performance when $\beta=0.5$ rather than when $\beta=0.7$ (with a larger $\beta$, $m$ is trained to handle more complex noise). We conjecture the main reason is that the noise in the training data when $\beta=0.7$ is beyond the capacity of $m$, which makes $m$ converge poorly during training. Overall, with $\beta$ set as 0.5 and $\gamma$ set as 0.3, the model achieves the best performance.


\begin{figure*}[t]
    \centering
    \includegraphics[scale=0.57]{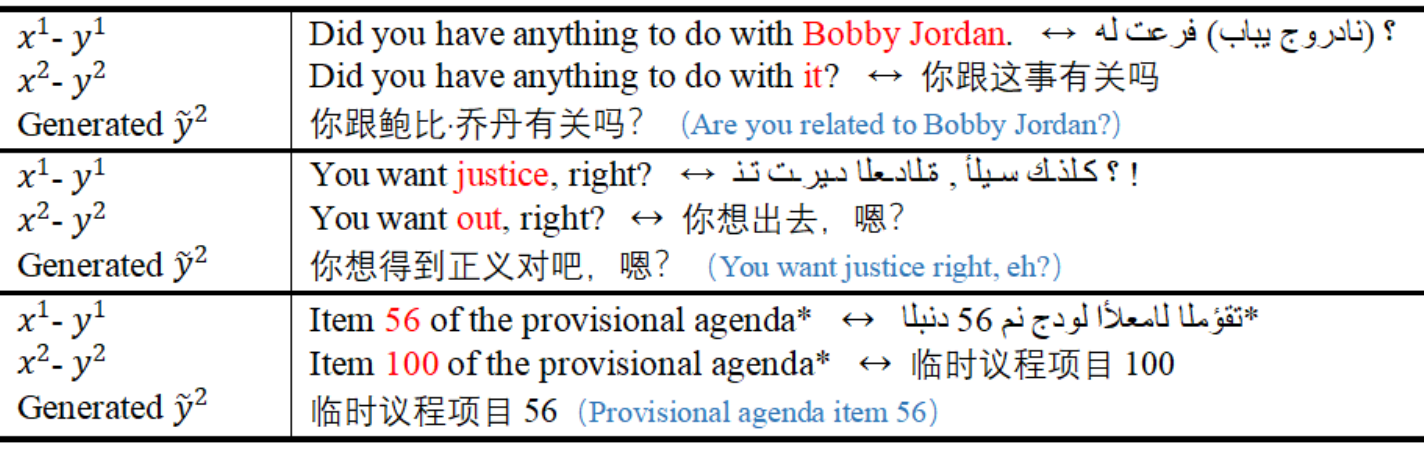}
    \caption {Examples constructed by EAG. "$x^1$-$y^1$" and "$x^2$-$y^2$" represent the bilingual examples in English $\rightarrow$ Arabic and English $\rightarrow$ Chinese respectively. $\tilde{y}^2$ is the generated Chinese sentence. The words in red color are the different word compositions between $x^1$ and $x^2$. The Google translations (in English) for $\tilde{y}^2$ are also provided.}
    \label{fig:study case}
    \vspace{-0.2cm}
\end{figure*}

\begin{table}[ht]
\centering
\scalebox{0.90}{
	\begin{tabular}{l|cc}
	\toprule[2pt]
	\# & system & BLEU \\
    \midrule[1pt]
        \textbf{0} & \textbf{EAG} & \textbf{17.9} \\
		1 & w/o $m$ & 15.7 \\
		2 & w/ $m$ as Transformer-big & 18.1\\
		3 & w/ $m$ as Transformer-deep & 17.8 \\
	\bottomrule[2pt]
	\end{tabular}}
\caption{\label{tab:ability_m} Results on the ability of $m$.}
\end{table}

\subsection{The ability of $m$}
We test how the ability of $m$ affects the final performance. Apart from the Transformer-base, i.e., the default setting used in EAG, we also test other two settings, namely Transformer-big \cite{vaswani2017attention} and Transformer-deep (20 encoder layers and 4 decoder layers). With different settings, $m$ is expected to perform different abilities in the generation process. The experimental results are presented in Table \ref{tab:ability_m}. We can find that if we remove $m$, the final performance drops dramatically (comparing \#0 with \#1). This shows that the generation step plays a significant role in the proposed EAG. However, by comparing among \#0, \#2 and \#3, we can find that the ability for $m$ shows little effect on the final performance. Taking all of \#0, \#1, \#2 and \#3 into consideration, we can reach a conclusion that, the generation step is very important for EAG and a simple generation model, i.e., a baseline NMT model, is enough to achieve strong performance.

\begin{table}[ht]
\centering
 \scalebox{0.90}{
	\begin{tabular}{l|cc}
	\toprule[2pt]
	System & w/o BT & w/ BT \\
    \midrule[1pt]
		extraction-based C-MNMT & 16.5  & 17.6 \\
		\textbf{EAG} & \textbf{17.9} &  \textbf{18.8} \\
	\bottomrule[2pt]
	\end{tabular}
	}
\caption{\label{tab:bt_result} Results on back-translation.  We report the average BLEU score on the non-English language pairs.}
\end{table}

\subsection{Back-translation}
We are very curious about how EAG works with back-translation (BT). To investigate this problem, we utilize the extraction-based C-MNMT model to decode the non-English monolingual sentences in the candidate aligned examples extracted by EAG, and then get the synthetic non-English sentence pairs by pairing the decoded results with the original sentences. The reversed sentence pairs are appended into the training corpus for the MNMT models. The experimental results are presented in Table \ref{tab:bt_result}. We find that BT improves the performances of both the two systems. Additionally, BT can work as a complementary to the proposed EAG.

\subsection{Case study and weaknesses}
Figure \ref{fig:study case} presents some examples constructed by EAG, each of which includes the extracted candidate aligned example and the generated sentence for Arabic $\rightarrow$ Chinese. The extracted candidate aligned example contains two bilingual examples, which are extracted from Arabic $\rightarrow$ English and Chinese $\rightarrow$ English respectively.  In Figure \ref{fig:study case}，the two bilingual examples in case one are extracted as a candidate aligned example as their English sentences have high similarity. And there is a composition gap between $x^1$ and $x^2$ since "Bobby Jordan" is mentioned in $x^1$, but not in $x^2$. By comparing the generated Chinese sentence $\tilde{y}^2$ with the original sentence $y^2$, we can find that $\tilde{y}^2$ is modified from $y^2$ by inserting the Chinese words `` 鲍比乔丹", which has the same meaning with "Bobby Jordan". Therefore, the generated $\tilde{y}^2$ is aligned to $x^1$ and $y^1$. In case 2, the Chinese word ``出去 (out)" in $y^2$ has been replaced with Chinese words "得到正义 (justice)" in  $\tilde{y}^2$, which makes the $\tilde{y}^2$ aligned to $x^1$ and $y^1$. Case 3 in Figure \ref{fig:study case} behaves similarly.

While achieving promising performance, the proposed approach still has some weaknesses in the real application: 1) The two-step pipeline performed by EAG is somewhat time-consuming compared to \citet{freitag2020complete}; 2) The generated multi-way aligned examples by EAG are sometimes not strictly aligned as the generation process does not always perform perfectly.

\section{Conclusion and Future work} \label{sec: conclusion}
In this paper, we propose a two-step approach, i.e., EAG, to construct large-scale and high-quality multi-way aligned corpus from English-centric bilingual data. To verify the effectiveness of the proposed method, we conduct extensive experiments on two publicly available corpora, WMT-5 and Opus-100. Experimental results show that the proposed method achieves substantial improvements over strong baselines consistently. There are three promising directions for the future work. Firstly, we plan to test whether EAG is applicable to the domain adaptation problem in NMT. Secondly, we are interested in applying EAG to other related tasks which need to align different example pairs. Finally, we want to investigate other model structures for the generation process.

\section*{Acknowledgments}
The authors would like to thank the anonymous reviewers of this paper, and the anonymous reviewers of the previous version for their valuable comments and suggestions to improve our work.

\bibliography{acl_latex}
\bibliographystyle{acl_natbib}

\newpage
\clearpage
\appendix

\section{Appendix}
\label{sec:appendix}
\maketitle



\subsection{Structure and training process for $m$}
\label{app:model}
We take transformer-big as the configuration for $m$. The word embedding dimension, head number, and dropout rate are set as 1024, 16, and 0.3 respectively. The model is trained on 8 V100 GPU cards, with learning rate, max-token, and update-freq set as 0.01, 8192, and 10 respectively. 

We train $m$ on the self-constructed examples with early-stopping. For the original example ($x^2, y^2$), we feed $m$ with the input format "$x^2$ <sep> $\tilde{y}^2$", where $\tilde{y}^2$ is the noised form of $y^2$, "<sep>" is a special token utilized to denote the sentence boundary. Similar to the traditional NMNT model, $m$ is trained to predict the original target sentence $y^2$.  

\subsection{Algorithm for our approach}
The algorithm for the proposed EAG is detailed as \ref{app:algorithm}.
\label{app:algorithm}

\setlength{\textfloatsep}{0.1cm}
\begin{algorithm*}[t!]
\setstretch{0.65}
\caption{Generating final aligned corpus: Given the aligned candidates $\{ x^1, y^1, x^2, y^2 \}$; an NMT model $m$, noisy probability $\beta$, word list $W_b$; return the final aligned corpus $\{{y}^1, \tilde{y}^2\}$}
\label{alg:training}
\begin{algorithmic}[1]
\Procedure{Noising}{$y^2_i$, $\beta$, $W_b$}
\State $\hat{y}^2_i \gets y^2_i$
\For {$t \in 0, \ldots, |\hat{y}^2_i|-1$}
  \State generate random float $\alpha \in (0,1)$
  \If {$\alpha < \beta$}
    \State perform insertion, removal or substitution on the position $t$ of $\hat{y}^2_i$ based on $W_b$
  \EndIf
\EndFor
\State \textbf{return} $\hat{y}_i$ 
\EndProcedure
\end{algorithmic}

\begin{algorithmic}[1]
\Procedure{Training}{$x^2$, $y^2$}
\State initialize $m$ randomly
\While{not convergence}
\For{$i \in 1, \ldots, |x^2|$}
  \State $\hat{y}^2_i \gets \textsc{Noising}(y^2_i, \beta$)
  \State train $m$ with the example $([x^2_i;\hat{y}^2_i],y^2_i)$ 
 \EndFor
 \EndWhile
 \State \textbf{return} $m$
\EndProcedure
\end{algorithmic}

\begin{algorithmic}[1]
\Procedure{Generating}{$x^1$, $x^2$, $y^1$, $y^2$, $m$}
\For{$i \in 1, \ldots, |x^1|$}
  \State get $\tilde{y}^2_i$ by performing the inference step of the well-trained $m$ with input $[x^1_i;y^2_i]$
  \State get the final aligned example by pairing $y^1_i$ with $\tilde{y}^2_i$
\EndFor
\State \textbf{return} $\{y^1, \tilde{y}^2\}$
\EndProcedure
\end{algorithmic}
\end{algorithm*}
\setlength{\floatsep}{0.1cm}

\subsection{Concrete BLEU score for each language pair}
\label{app:bleu}
\begin{table*}[ht]
\centering
 \begin{tabular}{l||c|c|a|c|c|c}
   & cs & de & en & es & fr & ru\\ \hline \hline
   cs &  & 17.1 & 30.8 & 21.2 & 22.4 & 14.3 \\ \hline
   de & 16.2 &  & 30.4 & 28.0 & 29.3 & 8.1 \\ \hline
   \rowcolor{Gray}
   en & 25.2 & 25.9 &  & 33.2 & 35.6 & 24.5 \\ \hline
   es & 16.3 & 23.1 & 35.4 &  & 35.1 & 20.0 \\ \hline
   fr & 15.2 & 22.7 & 33.5 & 33.0 &  & 18.7 \\ \hline
   ru & 13.2 & 7.3 & 29.0 & 23.5 & 25.0 &  \\
   \end{tabular}
  \caption{The BLEU score for the bilingual systems on WMT-5.}
  \label{tab:result for bilingual system}
\end{table*}

\begin{table*}[htb]
\centering
 \begin{tabular}{l||c|c|a|c|c|c}
   & cs & de & en & es & fr & ru\\ \hline \hline
   cs &  & 19.5 & 30.5 & 22.0 & 20.8 & 8.9 \\ \hline
   de & 6.1 &  & 31.4 & 17.5 & 21.0 & 4.0 \\ \hline
   \rowcolor{Gray}
   en & 24.2 & 27.1 &  & 33.2 & 34.4 & 24.1 \\ \hline
   es & 4.4 & 8.3 & 34.4 &  & 19.4 & 8.9 \\ \hline
   fr & 3.8 & 10.9 & 32.5 & 23.9 &  & 6.2 \\ \hline
   ru & 4.5 & 10.0 & 29.0 & 19.2 & 8.4 &  \\
   \end{tabular}
  \caption{The BLEU score for the standard MNMT on WMT-5.}
  \label{tab:result for MNMT system}
\end{table*}

\begin{table*}[htb]
\centering
 \begin{tabular}{l||c|c|a|c|c|c}
   & cs & de & en & es & fr & ru\\ \hline \hline
   cs &  & 22.8 & 30.8 & 27.1 & 29.2 & 22.3 \\ \hline
   de & 21.4 &  & 30.4 & 26.5 & 29.4 & 20.8 \\ \hline
   \rowcolor{Gray}
   en & 25.2 & 25.9 &  & 33.2 & 35.6 & 24.5 \\ \hline
   es & 23.1 & 23.0 & 35.4 &  & 32.6 & 22.9 \\ \hline
   fr & 21.7 & 22.8 & 33.5 & 30.5 &  & 22.0 \\ \hline
   ru & 21.6 & 20.4 & 29.0 & 27.1 & 28.0 &  \\
   \end{tabular}
  \caption{The BLEU score for the pivot system on WMT-5.}
  \label{tab:result for pivot system}
\end{table*}

\begin{table*}[htb]
\centering
 \begin{tabular}{l||c|c|a|c|c|c}
   & cs & de & en & es & fr & ru\\ \hline \hline
   cs &  & 24.9 & 30.9 & 29.3 & 31.2 & 26.0 \\ \hline
   de & 23.5 &  & 30.0 & 30.2 & 31.5 & 23.3 \\ \hline
   \rowcolor{Gray}
   en & 25.6 & 26.8 &  & 34.1 & 35.2 & 25.1 \\ \hline
   es & 24.2 & 25.4 & 35.0 &  & 35.3 & 24.6 \\ \hline
   fr & 22.9 & 24.9 & 34.0 & 32.5 &  & 23.3 \\ \hline
   ru & 24.2 & 22.6 & 29.7 & 28.6 & 29.8 &  \\
   \end{tabular}
  \caption{The BLEU score for the extraction-based C-MNMT on WMT-5.}
  \label{tab:result for extraction-based C-MNMT system}
\end{table*}

\begin{table*}[htb]
\centering
 \begin{tabular}{l||c|c|a|c|c|c}
   & cs & de & en & es & fr & ru\\ \hline \hline
   cs &  & 26.6 & 30.5 & 30.9 & 32.8 & 27.6 \\ \hline
   de & 25.0 &  & 30.1 & 30.1 & 32.5 & 24.0 \\ \hline
   \rowcolor{Gray}
   en & 25.1 & 26.9 &  & 34.7 & 35.5 & 25.6 \\ \hline
   es & 24.9 & 25.7 & 35.3 &  & 36.5 & 25.2 \\ \hline
   fr & 23.9 & 25.4 & 34.2 & 33.8 &  & 23.7 \\ \hline
   ru & 24.7 & 23.8 & 30.0 & 29.8 & 30.4 &  \\
   \end{tabular}
  \caption{The BLEU score for EAG on WMT-5.}
  \label{tab:result for EAG}
\end{table*}
In this section, we present the concrete BLEU score for each language pair on the corpus of WMT-5. Table \ref{tab:result for bilingual system}, \ref{tab:result for MNMT system}, \ref{tab:result for pivot system}, \ref{tab:result for extraction-based C-MNMT system}, and \ref{tab:result for EAG} show the translation performance for the bilingual system, standard MNMT system, pivot system, extraction-based system and the EAG respectively.


 \end{CJK} 
\end{document}